\documentclass[letterpaper, 10 pt, conference]{ieeeconf}  

\IEEEoverridecommandlockouts                              

\usepackage{amsmath} 
\usepackage{amssymb}  
\usepackage{graphicx}
\usepackage{multirow}
\usepackage{booktabs}
\usepackage{float}
\usepackage{graphicx}

\usepackage{tikz}
\usepackage{textcomp}
\usepackage{hyperref}
\usepackage{lipsum}

\usepackage{tabularx}
\usepackage{cite}

\newcommand\copyrighttext{%
  \footnotesize \textcopyright 2021 IEEE. Personal use of this material is permitted.
  Permission from IEEE must be obtained for all other uses, in any current or future
  media, including reprinting/republishing this material for advertising or promotional
  purposes, creating new collective works, for resale or redistribution to servers or
  lists, or reuse of any copyrighted component of this work in other works.
  DOI: \href{<http://tex.stackexchange.com>}{<DOI No.>}}
\newcommand\copyrightnotice{%
\begin{tikzpicture}[remember picture,overlay]
\node[anchor=south,yshift=10pt] at (current page.south) {\fbox{\parbox{\dimexpr\textwidth-\fboxsep-\fboxrule\relax}{\copyrighttext}}};
\end{tikzpicture}%
}

\graphicspath{ {Figures/} } 
\title{\LARGE \bf
Gait-based Frailty Assessment using Image Representation of IMU Signals and Deep CNN *
}

\author{Muhammad Zeeshan Arshad$^{1}$, Dawoon Jung$^{1}$, Mina Park$^{1}$, Hyungeun Shin$^{2}$, 
\\Jinwook Kim$^{1}$, and Kyung-Ryoul Mun$^{1}$
\thanks{*This research was supported by the Korea Institute of Science and Technology Institutional Program (Project No. 2E31051) and in part by the High-Tech Based National Athletic Performance Improvement (Winter) Project from the Korea Sports Promotion Foundation.}
\thanks{$^{1}$Muhammad Zeeshan Arshad, Dawoon Jung, Mina Park, Jinwook Kim, and Kyung-Ryoul Mun are with the Center for Artificial Intelligence, KIST, Seoul, Republic of Korea.
        {\tt\small krmoon02@kist.re.kr}}%
\thanks{$^{2}$Hyungeun Shin is with Department of Biomedical Science and Technology, Graduate School, Kyung Hee University, Seoul, Republic of Korea.}
}
\begin{document}

\maketitle
\copyrightnotice
\thispagestyle{empty}
\pagestyle{empty}

\begin{abstract}

Frailty is a common and critical condition in elderly adults, which may lead to further deterioration of health. However, difficulties and complexities exist in traditional frailty assessments based on activity-related questionnaires. These can be overcome by monitoring the effects of frailty on the gait. In this paper, it is shown that by encoding gait signals as images, deep learning-based models can be utilized for the classification of gait type. Two deep learning models (a) SS-CNN, based on single stride input images, and (b) MS-CNN, based on 3 consecutive strides were proposed. It was shown that MS-CNN performs best with an accuracy of 85.1\%, while SS-CNN achieved an accuracy of 77.3\%. This is because MS-CNN can observe more features corresponding to stride-to-stride variations which is one of the key symptoms of frailty. Gait signals were encoded as images using STFT, CWT, and GAF. While the MS-CNN model using GAF images achieved the best overall accuracy and precision, CWT has a slightly better recall. This study demonstrates how image encoded gait data can be used to exploit the full potential of deep learning CNN models for the assessment of frailty.

\end{abstract}

\section{INTRODUCTION}
Advancements in healthcare and medical technologies in the last few decades have dramatically changed the world demographics. According to a recent UN report, our society is aging so rapidly that by the year 2050 the number of older persons worldwide would double to over 1.5 billion compared to the year 2020 \cite{nations2020world}. This continuous trend has put forward new challenges to sustain healthcare and eldercare which must be dealt with major focus shift in healthcare services towards early diagnosis and preventive interventions and strategies.

Frailty, a condition that prevails in old age, is considered as a state of increased vulnerability or a precursor to more adverse outcomes like morbidity, falls, institutionalization, disabilities, and mortality \cite{frieswijk2004effect, campbell1997unstable, van2008iana, rockwood2007frailty}. Frailty can be defined as a gradual decrease in physiological and functional reserves as well as resistance to internal or external stressors. \cite{campbell1997unstable,rockwood1994frailty,clegg2013frailty}. However, if identified and treated at an early stage, the onset of frailty and the consequent effects could be delayed and in some cases avoided altogether for example through early implementation of fall prevention strategies \cite{morley2008diabetes} and multicomponent exercises (i.e. strength, endurance, flexibility, and balance training) focused on improving the functional capacity of the elderly \cite{villareal2011regular}. These would help to reduce the serious personal and societal impact that comes with falling, resulting injuries, and their healthcare costs.

One of the most common methods to assess frailty in a clinical setting is the "Fried Criteria" \cite{fried2001frailty} which identified five frailty phenotypes, namely, shrinking, weakness, slowness, exhaustion, and low activity. The other widely used criteria include the Study of Osteoporotic Fractures (SOF) scale \cite{ensrud2008comparison} and the FRAIL scale \cite{van2008frailty}. However, these criteria rely on answers from the patient to activity-related questionnaires. Trained professionals are required to explain the questions, measure different phenotypes, and assess the answers carefully to overcome subjective interpretation and judgments. Furthermore, medical history may be essential for more accurate assessment in some cases. All these factors make these frailty assessments too cumbersome to use in the acute hospital environment.

Gait has been strongly linked with frailty since the decline in mobility and balance is one of the major indicators of frailty and can be assessed through gait analysis \cite{schwenk2014frailty}. Gait velocity \cite{gill2001two, rothman2008prognostic} and variability \cite{montero2011gait} are among the most significantly affected gait parameters in frail patients. The use of wearable inertial sensors \cite{schwenk2015wearable, greene2014frailty, godfrey2017wearables, razjouyan2018wearable} to assess the gait not only simplifies and speeds up the assessment procedure but also gives a more objective evaluation of frailty status.

In this paper we propose a frailty assessment method which uses deep learning CNN models with image representation of gait data collected through  Inertial Measurement Unit (IMU) sensors. We propose two CNN models based on single and multi-stride data. We also compare different imaging techniques to evaluate their suitability for gait data. 

 \begin{figure}[t!]
      \centering
      \includegraphics[width=\linewidth]{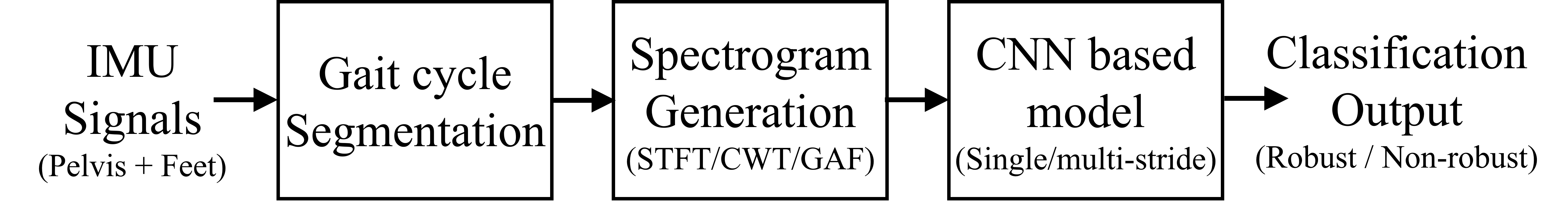}
      \caption{Flow chart of proposed frailty classification methodology}
      \label{MethodFlow}
\end{figure} 

\section{Methodology}

The frailty assessment methodology proposed in this paper starts with the acquisition of gait data using IMU sensors as shown in Fig. \ref{MethodFlow}. Then gait cycle segmentation is performed to extract individual gait strides. These segmented signals are then encoded into images using different image encoding techniques. Finally, the encoded images are fed into CNN-based models for classifying the gait type. The following subsections describe the proposed methodology steps.

\subsection{Participants and Data collection} The data for this study was collected from 71 elderly subjects. Inclusion criteria were being older, community-dwelling adults with age $\geq$ 65 years, and having the ability to walk independently without a walking aid. Subjects with any musculoskeletal abnormalities or severe gait and balance disorders that may limit their ability to walk at least 20m were excluded. The demographic characteristics of the cohort are presented in Table \ref{table_Demographics}. A brief introduction about the study was provided to the participants. A written and signed informed consent was taken from all participants before the experiment. This study was approved by the Institutional Review Board of Kyung Hee University Medical Center.

\begin{table}[hbt!]
\centering
\caption{Demographic information of the patients}
\label{table_Demographics}
\begin{tabular}{@{}ll@{}}
\toprule
\textbf{Variable}    & \textbf{Mean $\pm$ SD (Range)}              \\ \midrule
Age (years) & 77.56 $\pm$ 3.92 (71-86)       \\
Height (cm) & 155.90 $\pm$ 18.35 (141.2-175) \\
Weight (kg) & 60.08 $\pm$ 9.12 (41.4-93)     \\ \midrule
Gender      &                                \\
\quad - Male (\%)   & 38.02                          \\
\quad - Female (\%) & 61.98                          \\ \bottomrule
\end{tabular}
\end{table}

The frailty assessment of the subjects was made using the FRAIL scale \cite{van2008frailty} which assigns scores from 0 to 5 in an increasing order of frailty severity. A score of 0 implies robust, 1-2 implies pre-frail, and 3–5 implies frail status. However, the last two can be combined to be referred as a non-robust state. Therefore, for this work, subjects were divided into two groups, robust (with a score of zero) and non-robust (with a score exceeding zero), in terms of their frailty status. Out of 71 subjects, 26 were termed as robust and the rest of the 45 as non-robust.

 \begin{figure}[hbt!]
      \centering
      \includegraphics[width=0.95\columnwidth]{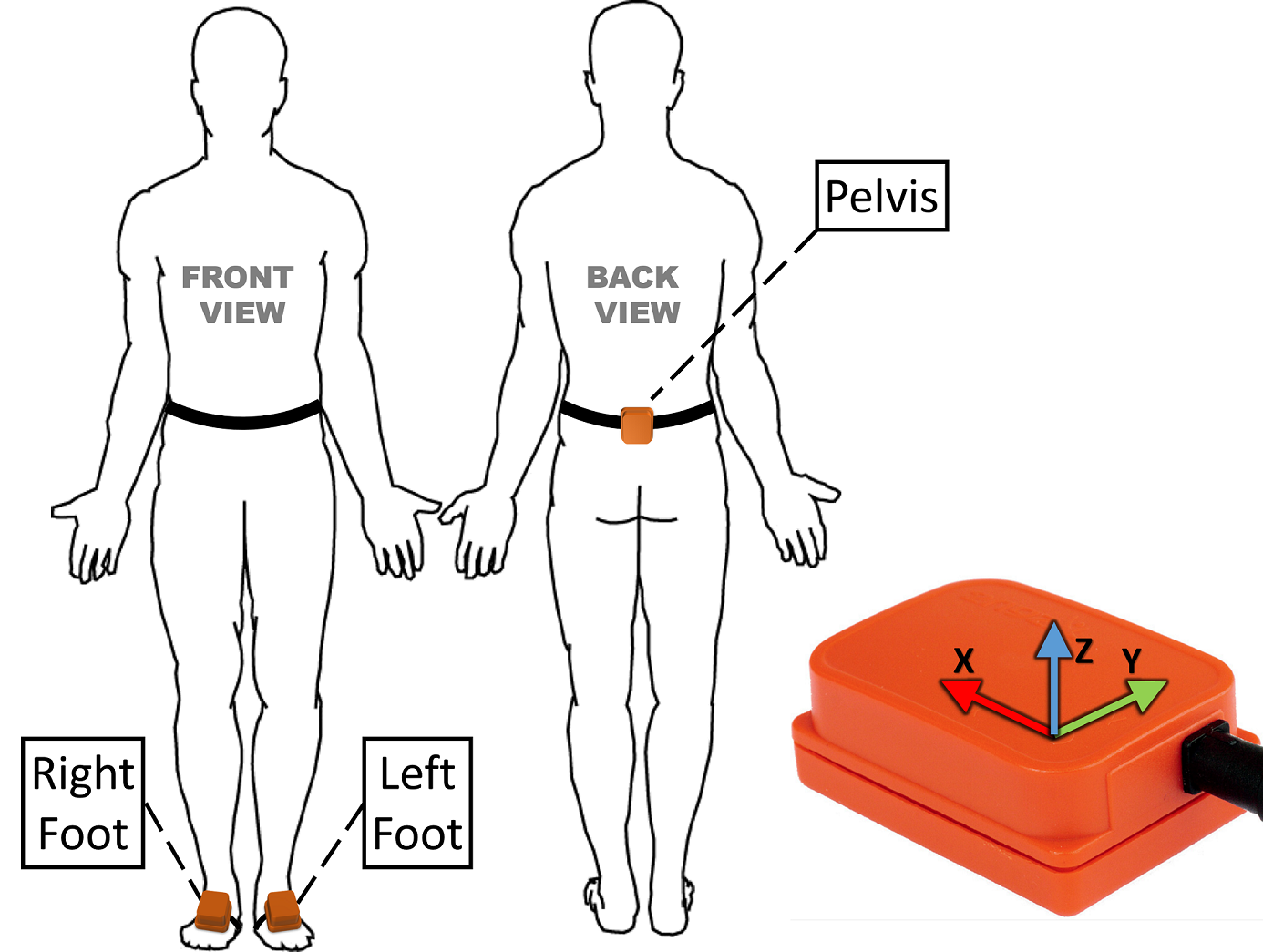}
      \caption{IMU sensor and locations  }
      \label{Sensor}
\end{figure} 

Three IMU sensors (Xsens MVN, Enschede, Netherland) were mounted on the subjects' body as shown in Fig. \ref{Sensor}. One sensor was mounted on the posterior pelvis (positioned flat on the sacrum) and one on each foot (positioned at the middle of the bridge of the foot). The participants were instructed to walk 20m while each IMU sensor recorded the acceleration and angular velocity in its own 3D local coordinate system at a sampling rate of 100Hz. All sensor signals were passed through a noise filtering, 0.1-15Hz band-pass filter before being upsampled to 1000Hz through cubic spline interpolation. The data was then normalized to the range [-1,1].

 \begin{figure}[hbt!]
      \centering
      \includegraphics[width=\linewidth]{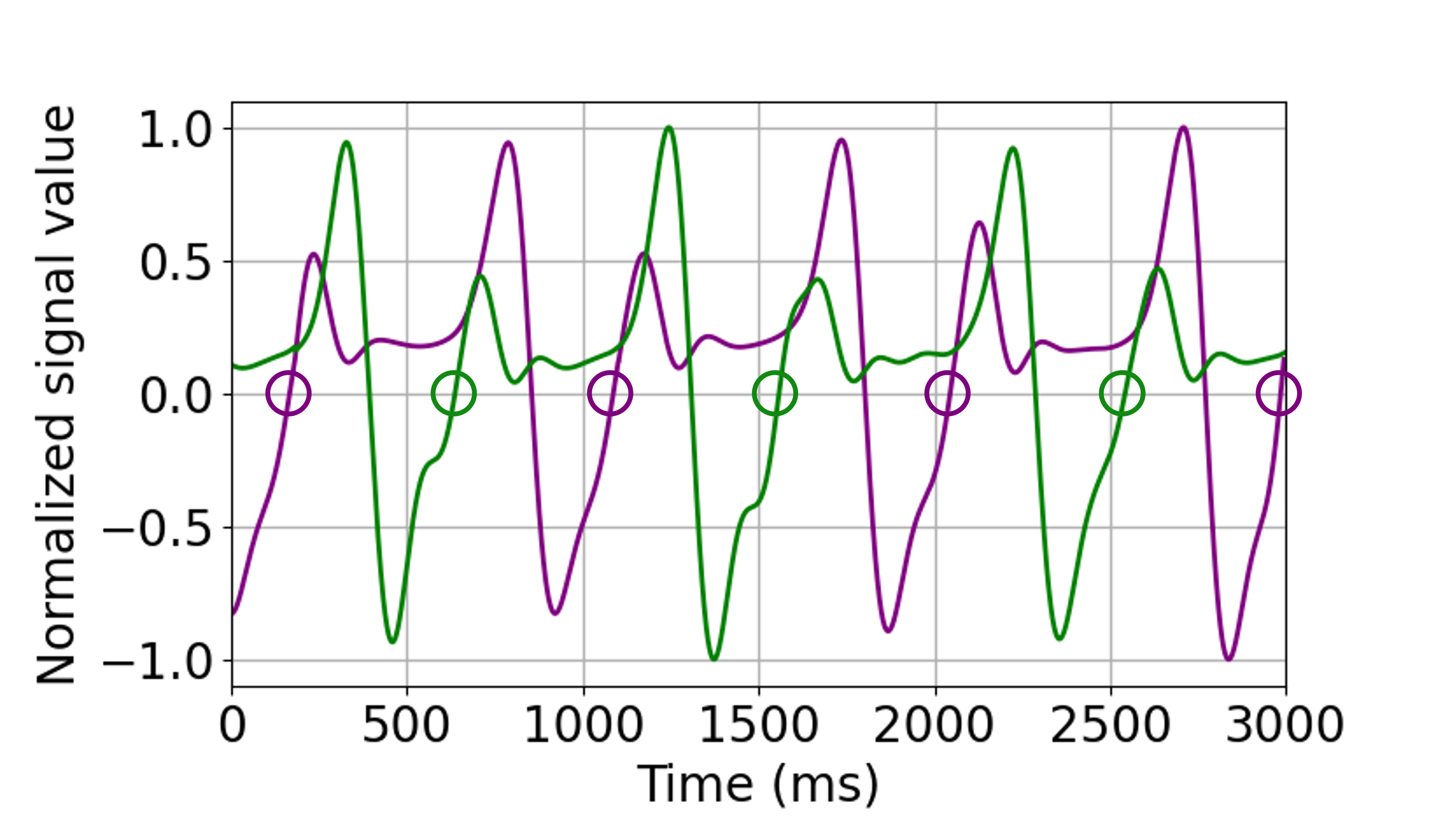}
      \caption{An example of gait heel-strike events (circled) detected through the Angular Velocity Y signal of right foot (green line) and left foot (indigo line)}
      \label{GyroscopeSignal}
\end{figure} 

All six acceleration and gyroscope values from x-, y-, and z-axes were used for the IMU sensor located at the pelvis. For each foot, however, the x-, and y-axis values were combined to get the magnitude of the resultant acceleration and angular velocity. This enables to overcome errors that arise from slight variations in sensor orientation at the bridge of the foot. Hence, a total of 14 features (6 from the pelvis and 4 from each foot) were extracted from the IMU sensors. The gait segmentation was performed through detection of heel-strike (HS) events from the angular velocity (sagittal plane) as shown in Fig \ref{GyroscopeSignal}. The resulting individual stride signals for each subject were then used for spectrogram generation. A total of 642 strides were extracted, consisting of 205 strides from the robust group and 437 strides from the non-robust group. 

\subsection{Image Generation} By converting signals to images, the capabilities of CNN were exploited to extract more diverse features from complex local patterns in the images.

Short-time Fourier transform (STFT) was used to get frequency and amplitude of localized waves within a span of temporal window \cite{grochenig2001foundations} as follows:

\begin{equation}
\label{eqn_1}
STFT\left \{ x(t) \right \}X(\tau, \omega) ={\intop_{ -\infty }^{+ \infty }   x(t)\omega(t - \tau) e^{-j \omega t}  dt}
\end{equation}

where $x(t)$, $\omega(t)$, and $\tau$ denote the signals to be transformed, a window function (Gaussian window in this study) centered around zero, and a time shift, respectively. The spectrogram is generated by squaring the STFT magnitude as follows:

\begin{equation}
\label{eqn_example}
spectrogram\{x(t)\} = \|X(\tau, \omega)\|^{2}
\end{equation}

\begin{figure}[hbt!]
      \centering
      \includegraphics[width=0.95\columnwidth]{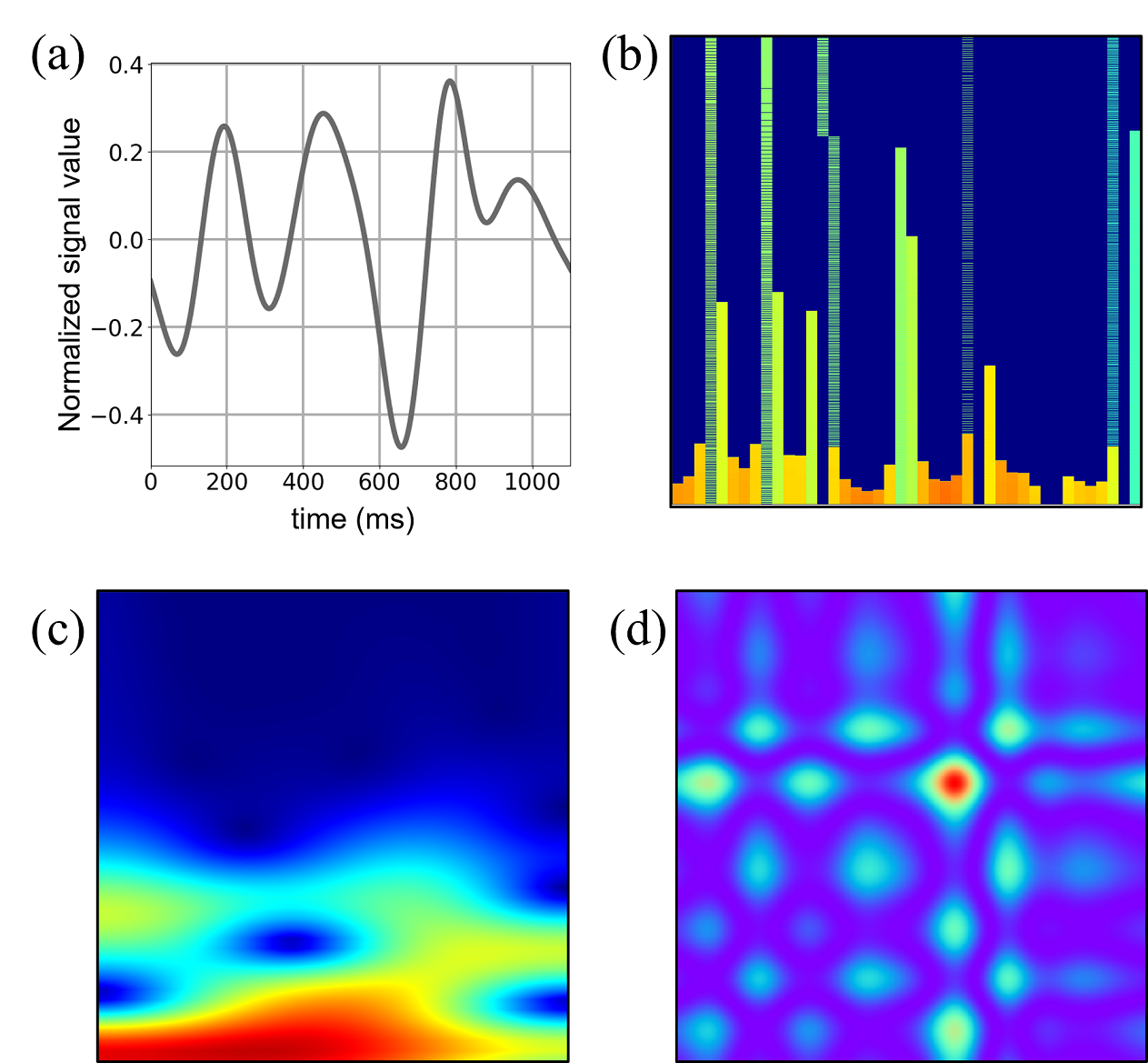}
      \caption{Example images generated for accelerometer x-signal on pelvis sensor (a) using STFT (b), CWT (c), and GAF (d)}
      \label{Spectrograms}
\end{figure}

The frequency range was restricted to 6Hz and the time resolution was fixed at 0.1s. Figure \ref{Spectrograms} (a) shows the STFT spectrogram for the x-axis pelvis acceleration signal. The brighter colors correspond to a higher energy frequency component.

Compared to Fourier transform where signals are decomposed into sinusoids of different frequencies, in the Continuous Wavelet Transform (CWT), the signals are decomposed into shifted or scaled shapes from the mother wavelet. CWT is defined \cite{mallat1999wavelet} as:

\begin{equation}
\begin{aligned}
  C(a,b) =  {\frac{1}{ \sqrt{ \begin{vmatrix}a\end{vmatrix}}}} {\intop_{ -\infty }^{+ \infty }  s(t) \it \Psi^\ast \begin{pmatrix}\frac{t-b}{a}\end{pmatrix}dt} \\
  a \in R^{+} - \{0\}, b \in R, 
\end{aligned}
\end{equation}

where $f(t)$ is the input signal and $\Psi_{a,b}(t)$ is the mother wavelet with 'a' as the scale factor and 'b' as the shift factor. The CWT for this study used the Morse wavelet with 20 voices per octave and a frequency range restricted to 15Hz. The visual representation of the CWT of a signal is referred to as a scalogram. The scalogram for the x-axis pelvis acceleration signal is presented in Fig. \ref{Spectrograms} (b).

Granular Angular Field (GAF) is a method which uses polar coordinates based matrix to encode a time series into an image while still preserving the absolute temporal relationships in the series \cite{wang2015imaging}. In this study we use Gramian Angular Summation Field (GASF) which uses the cosine of the summation of angles from the polar coordinates and is given as:

\begin{equation}
\label{eqn_example_2}
\mathbf{GASF} = \left(
\begin{array}{ccc}
\cos(  \phi _{1} + \phi _{1} ) & \ldots & \cos(  \phi _{1} + \phi _{n} ) \\
\cos(  \phi _{2} + \phi _{1} ) & \ldots & \cos(  \phi _{2} + \phi _{n} ) \\
\vdots & \ddots & \vdots \\
\cos(  \phi _{n} + \phi _{1} ) & \ldots & \cos(  \phi _{n} + \phi _{n} )
\end{array} \right)
\end{equation}
Figure \ref{Spectrograms}  (c) shows the GASF for an example signal. All images from STFT, CWT, and GAF were produced with the dimensions 224x224 pixels. A total of 17,976 images were produced from the three sensors from the 71 participants.

 \begin{figure}[hbt!]
      \centering
      \includegraphics[width=0.8\columnwidth]{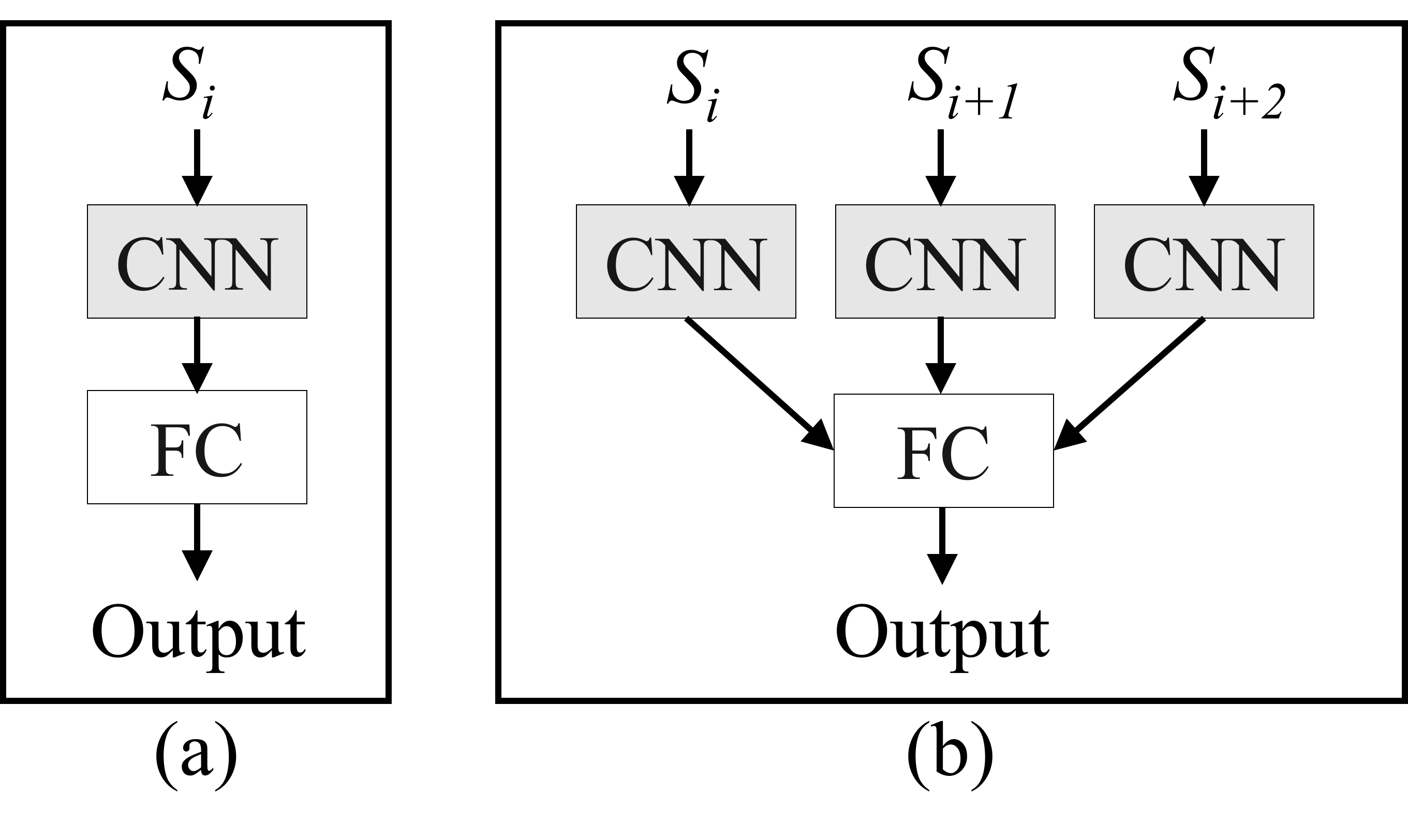}
      \caption{CNN model architecture (a) Single-stride-CNN (b) Multi-stride-CNN  }
      \label{ModelArc}
\end{figure} 

\subsection{Convolution Neural Network Modeling} To prepare the images for input to the CNN model, all images, sized 224 x 224 x 3 (3-channel RGB image) each, from each of the 14 parameters were stacked in the order of pelvis, right foot, and left foot. This way, for each stride, the input image set of size 224 x 224 x 42  was obtained. Two types of CNN models were proposed in this study: the Single-stride-CNN (SS-CNN) and Multi-stride-CNN (MS-CNN). SS-CNN uses a single stride image set ($S_i$) as input to train and classify between the two output classes robust and non-robust for training and testing. as shown in \ref{ModelArc} (a). However, MS-CNN uses three consecutive strides ($S_i, S_{i+1}, S_{i+2})$) of the subject for the task as in Fig. \ref{ModelArc} (b). Leaping one step ahead of just learning on features from each stride, the MS-CNN further enables identification and learning of features related to stride-to-stride variability in the gait. For $n$ consecutive strides, $n-2$ three-stride sets were extracted. For instance, for a subject having 10 consecutive strides in the data, 8 three-stride sets were obtained as [1, 2, 3], [2, 3, 4], [3, 4, 5] ... [8, 9, 10]. This resulted in a total of 360 stride sets, 103 from the robust group and 257 from the non-robust group.

\begin{figure*}
  \includegraphics[width=\textwidth,height=4cm]{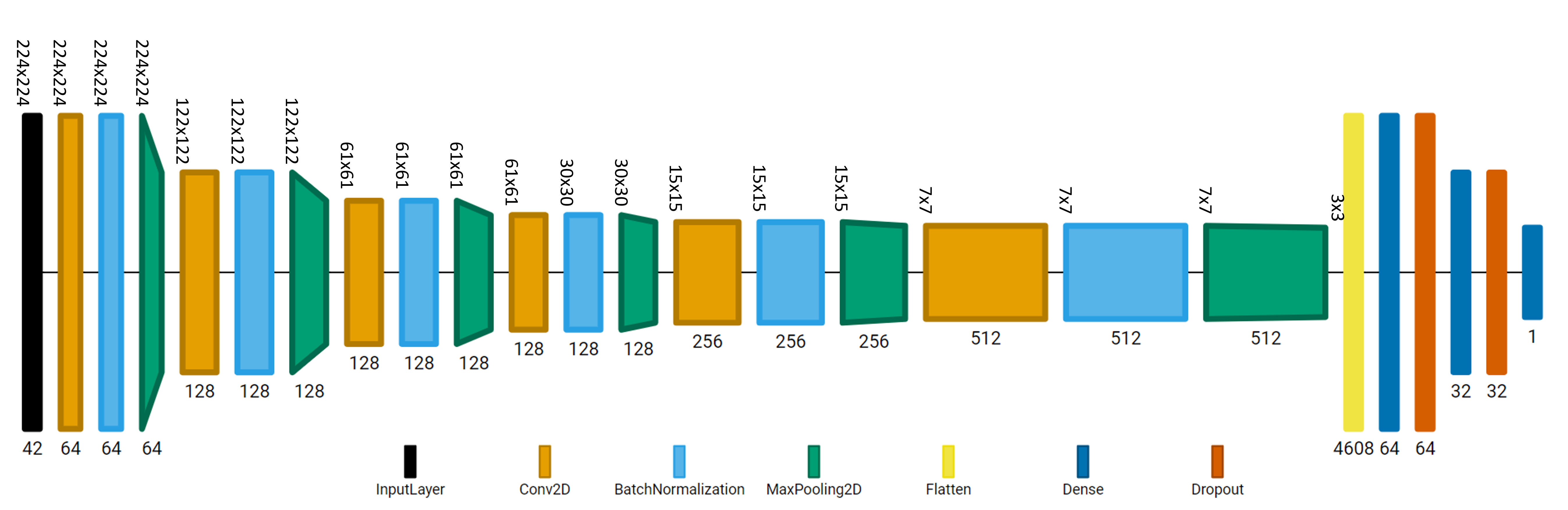}
  \caption{The detailed architecture of  CNN and Dense block in the CNN model}
  \label{ModelArc_detail}
\end{figure*}

The proposed architecture of SS-CNN consists of six convolutional blocks, followed by three fully connected layers and a sigmoid layer at the output as shown in Fig. \ref{ModelArc} (a). Each convolutional block contains a convolutional layer, a Rectified Linear Unit (ReLU) layer, a batch normalization (BN) layer, and a Maxpooling layer. The six convolutional layers have filter sizes 64, 128, 128, 128, 256, and 512. The kernel size is set to 20x20 for all. The three fully connected layers have lengths 64, 32, and 1 as the output. Dropout layers are applied to the fully connected layers. The detailed architecture of CNN and fully connected layers is given in Fig. \ref{ModelArc_detail}.

In MS-CNN, instead of a single stride image set, the input consisting of three consecutive stride image sets goes into separate convolutional blocks which are concatenated before the fully connected layers as shown in Fig. \ref{ModelArc} (b). Here, the design parameters for the convolutional blocks and the fully connected layers have been kept the same as in SS-CNN.

\subsection{Training and Testing} For both types of models, 90\% of images were selected for training and 10\% for testing, in a stratified manner as to keep the same class distribution as in the original data. A stratified shuffle split cross-validation with five folds was used to evaluate performance. The networks were trained using Adaptive learning rate (ADADELTA) to make the best of both, learning rate annealing and momentum training to converge faster. Initial values of the weights and bias were initialized with Glorot uniform initializer. An early stopping criterion was employed to stop training when the validation accuracy did not improve for 5 epochs. 

\begin{table}[hbt!]
\caption{Model results}
\label{table_Results}
\resizebox{\columnwidth}{!}{
\renewcommand{\arraystretch}{1.5} 
\begin{tabular}{lccccc}
\hline
\textbf{Model}                                                             & \multicolumn{1}{l}{\textbf{Image}} & \textbf{Accuracy} & \multicolumn{1}{l}{\textbf{Precision}} & \multicolumn{1}{l}{\textbf{Recall}} & \multicolumn{1}{l}{\textbf{F-Measure}} \\ \hline
\multirow{3}{*}{\begin{tabular}[c]{@{}l@{}}SS-\\ CNN\end{tabular}} & STFT                                & 0.66              & 0.722                                   & 0.843                                   & 0.778                                  \\
                                                                           & CWT                                 & 0.713              & 0.808                                   & 0.813                                    & 0.811                                  \\
                                                                           & GAF                                 & 0.773              & 0.810                                   & 0.878                                   & 0.843                                  \\ \hline
\multirow{3}{*}{\begin{tabular}[c]{@{}l@{}}MS-\\ CNN\end{tabular}}  & STFT                                & 0.72              & 0.808                                   & 0.889                                   & 0.847                                  \\
                                                                           & CWT                                 & 0.823              & 0.875                                   & 0.904                                  & 0.889                                   \\
                                                                           & GAF                                 & 0.851              & 0.912                                   & 0.896                                   & 0.904                                  \\ \hline
\end{tabular}
}
\end{table}

\section{RESULTS AND DISCUSSION} 
Table \ref{table_Results} shows the classification results for the two proposed models. As SS-CNN uses single stride images, hence the number of training samples was equal to the total number of strides available in the data. On the other hand, the MS-CNN requires a set of three consecutive strides, which reduces the number of training samples by about half. Nonetheless, the MS-CNN attains better performance compared to the SS-CNN for all three types of images. That is because the output of MS-CNN architecture relies not only on the features of each stride but more importantly on the gait dynamics and stride-to-stride variation within the three strides. This gives MS-CNN a considerable advantage over the SS-CNN, and hence it attains an average 8.2\% higher accuracy.

Models trained on GAF images achieved the best accuracy among the three, while CWT and STFT come next in sequence. Unlike CWT and STFT which are classical time-frequency methods, GAF is used to encode time series as images, while preserving the temporal correlations existing in the signal. The deep learning model uses this temporal dependence within the gait time series for a better prediction. For the SS-CNN model, GAF achieved an accuracy of 77.3\% while CWT achieved 71.3\% and STFT achieved 66\%. The sequence is followed in terms of precision, recall, and F-Measure for the SS-CNN. Moreover, for the MS-CNN model, GAF achieves an accuracy of 85.1\%, the best overall in this study, while CWT achieves 82.3\% and STFT achieves 72\%. Although MS-CNN also has GAF as the best model in terms of accuracy and precision, however, CWT has a slightly higher recall in this case. 

 \begin{figure}[hbt!]
      \centering
      \includegraphics[width=0.55\columnwidth]{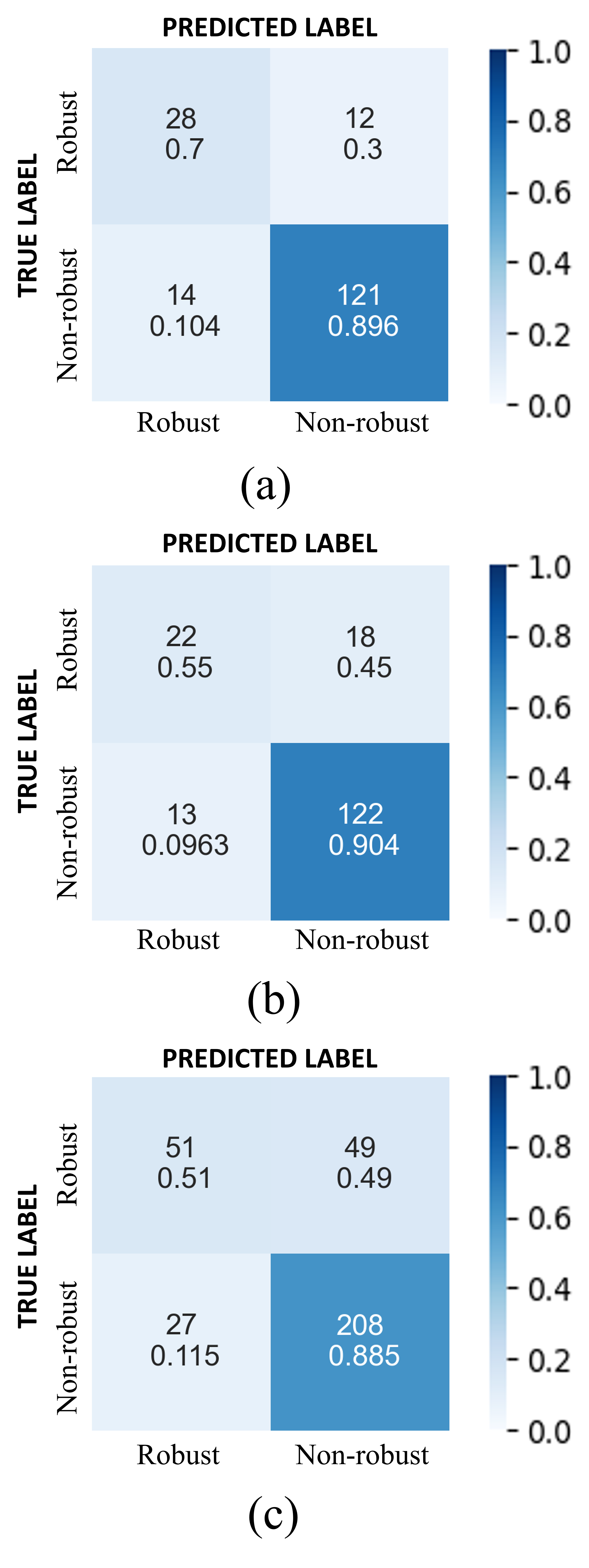}
      \caption{Confusion matrices for (a) MS-CNN using GAF (b) MS-CNN using CWT, and (c) SS-CNN using GAF}
      \label{ConfusionMatrix}
\end{figure} 

Figure \ref{ConfusionMatrix} shows the confusion matrices for the two best MS-CNN models and the best SS-CNN model. Comparing the two confusion matrices of the MS-CNN model with GAF input (Fig. \ref{ConfusionMatrix}(a)) and MS-CNN model with CWT input  (Fig. \ref{ConfusionMatrix}(b)) that GAF generated 12 false positives, much less than 18 in CWT's case, indicating the better precision achieved through GAF. However, CWT generated one less false negative and one more true positive compared to GAF, showing its slightly better recall performance. 

In Fig. \ref{ConfusionMatrix} (c), the confusion matrix for SS-CNN with GAF as input is given. The numbers are not directly comparable with those in MS-CNN because of the different number of samples, but we see from the normalized values that SS-CNN using GAF input gives 19\%  more false positives and over 1\% more false negatives compared to MS-CNN using GAF. This shows that its precision performance is much worse than its recall performance. 

While a better recall is more desirable than precision in disease classification, however in this case the recall performance improvement of MS-CNN using CWT over GAF is not significant ($<$1\%), however, it gives much lower precision performance compared to GAF (difference of 3.7\%). Therefore, we could say that overall, MS-CNN using GAF input provides a better assessment of frailty with its higher accuracy and F-measure.

Regarding the poorer performance of STFT compared to the CWT, STFT uses a fixed window length resulting in a fixed frequency resolution. Shorter window lengths give higher time resolution but deteriorate frequency resolution. Contrarily, longer window lengths give higher frequency resolution but reduce time resolution \cite{gabor1946theory}. This limits STFT's ability to perceive all time-frequency variations in the signal. This missing information fails to appear on the produced STFT spectrogram images, essentially causing a drop in its performance. CWT, on the other hand, works with variable window lengths, using shorter window lengths for higher frequencies and longer window lengths for lower frequencies \cite{rioul1991wavelets}. Hence, it gives a superior frequency-time resolution.

\section{CONCLUSIONS}
 Gait IMU data was collected for frail and non-frail elderly. It was shown that by encoding gait signals as images, deep learning-based models can be utilized for the classification of gait type. The proposed multi-stride-based CNN deep learning model using GAF as input achieved an accuracy of 85\% and precision of 0.912. This work shows that frailty can be assessed with high accuracy using gait IMU data encoded as images. Deep learning models can learn the features capturing the changes in temporal signal patterns and stride-to-stride variations. As future work, it will be interesting to explore the recurrent neural network-based deep learning models that are better suited for the identification of patterns in time-series data.       

\IEEEtriggeratref{22}

\bibliographystyle{IEEEtran}
\bibliography{IEEEabrv,IEEEexample}
\end{document}